\documentclass[10pt,twocolumn,letterpaper]{article}

\usepackage[pagenumbers]{cvpr} % To force page numbers, e.g. for an arXiv version

\usepackage{tabu}
\usepackage{tabularx}
\usepackage{lipsum}
\usepackage{times}
\usepackage{epsfig}
\usepackage{graphicx}
\usepackage{amsmath}
\usepackage{amssymb}
\usepackage{caption}
\usepackage{multirow}
\usepackage{float}
\usepackage{wrapfig}
\usepackage{nicefrac}
\usepackage[normalem]{ulem}
\usepackage[export]{adjustbox}
\usepackage{booktabs}

\usepackage{graphicx}
\usepackage{isomath}
\usepackage{tipa}
\usepackage{caption}
\usepackage{subcaption}
\usepackage{float}
\usepackage{gensymb}

\usepackage[pagebackref,breaklinks,colorlinks]{hyperref}

\usepackage[capitalize]{cleveref}
\crefname{section}{Sec.}{Secs.}
\Crefname{section}{Section}{Sections}
\Crefname{table}{Table}{Tables}
\crefname{table}{Tab.}{Tabs.}

\def\cvprPaperID{*****} % *** Enter the CVPR Paper ID here
\def\confName{CVPR}
\def\confYear{2022}

\begin{document}

\title{An Empirical Study of Self-supervised Learning Approaches \\ for Object Detection with Transformers}

% \author{%
%   Abhishek Singh Gehlot\thanks{Equal contribution. Listing order is alphabetical.} \quad
%   Gokul Karthik Kumar\footnotemark[1] \hspace{2} \quad
%   Sahal Shaji Mullappilly\footnotemark[1] \hspace{2}\\
%   Mohamed Bin Zayed University of Artificial Intelligence (MBZUAI) \\
%   Abu Dhabi, UAE\\
%   \texttt{\{abhishek.gehlot, gokul.kumar, sahal.mullappilly\}@mbzuai.ac.ae} \\}
  
\author{%
  Gokul Karthik Kumar \hspace{2pt} \quad
  Sahal Shaji Mullappilly \hspace{2pt} \quad
  Abhishek Singh Gehlot \\
  Mohamed Bin Zayed University of Artificial Intelligence (MBZUAI) \\
  Abu Dhabi, UAE\\
  \texttt{\{gokul.kumar, sahal.mullappilly, abhishek.gehlot\}@mbzuai.ac.ae} \\}
  
\maketitle
%%%%%%%%% ABSTRACT
\begin{abstract}
Self-supervised learning (SSL) methods such as masked language modeling have shown massive performance gains by pretraining transformer models for a variety of natural language processing tasks. The follow-up research adapted similar methods like masked image modeling in vision transformer and demonstrated improvements in the image classification task. Such simple self-supervised methods are not exhaustively studied for object detection transformers (DETR, Deformable DETR) as their transformer encoder modules take input in the convolutional neural network (CNN) extracted feature space rather than the image space as in general vision transformers. However, the CNN feature maps still maintain the spatial relationship and we utilize this property to design self-supervised learning approaches to train the encoder of object detection transformers in pretraining and multi-task learning settings. We explore common self-supervised methods based on image reconstruction, masked image modeling and jigsaw. Preliminary experiments in the iSAID dataset demonstrate faster convergence of DETR in the initial epochs in both pretraining and multi-task learning settings; nonetheless, similar improvement is not observed in the case of multi-task learning with Deformable DETR. The code for our experiments with DETR and Deformable DETR are available at \url{https://github.com/gokulkarthik/detr} and \url{https://github.com/gokulkarthik/Deformable-DETR} respectively.
\end{abstract}

%%%%%%%%% BODY TEXT
\section{Introduction:}
\label{sec:intro}

\begin{figure}[]
\centering
   \includegraphics[width=1\linewidth]{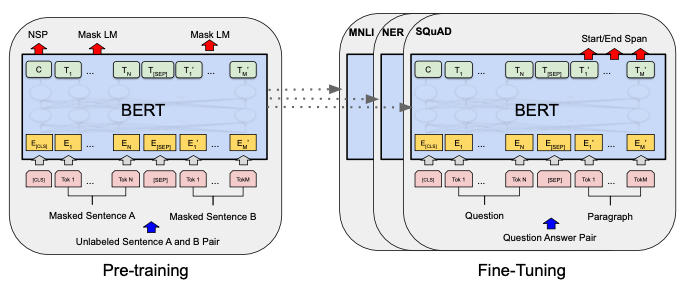}
   \caption{Overall pre-training and fine-tuning procedures for BERT from \cite{devlin2018bert}. Apart from output layers, the same architectures are used in both pre-training and fine-tuning. The same pre-trained model parameters are used to initialize models for different downstream tasks. During fine-tuning, all parameters are fine-tuned.}
   \label{fig:bert}
\end{figure}

The advent of Bidirectional Encoder Representations from Transformers (BERT) \cite{devlin2018bert} made the transformer architecture widely popular in the Natural Language Processing (NLP) community because of its superior performance. The work utilized huge amounts of unlabelled text corpus with the self-supervised learning tasks of Masked Language modeling (MLM) and Next Sentence Prediction (NSP) to pretrain the transformer network as shown in Figure \ref{fig:bert}. The learning objective of the Masked Language modeling is to predict the randomly masked words in a sentence whereas in Next Sentence Prediction the task is to identify whether the two input sentences are consecutive or not. Experiments on the benchmark datasets for a variety of downstream tasks such as Natural Language Inference, Named Entity Recognition and Question Answering demonstrated the outstanding performance of the task-specifically self-supervisingly pretrained transformer network when fine-tuned on task specific datasets.

\begin{figure}[]
\centering
   \includegraphics[width=0.9\linewidth]{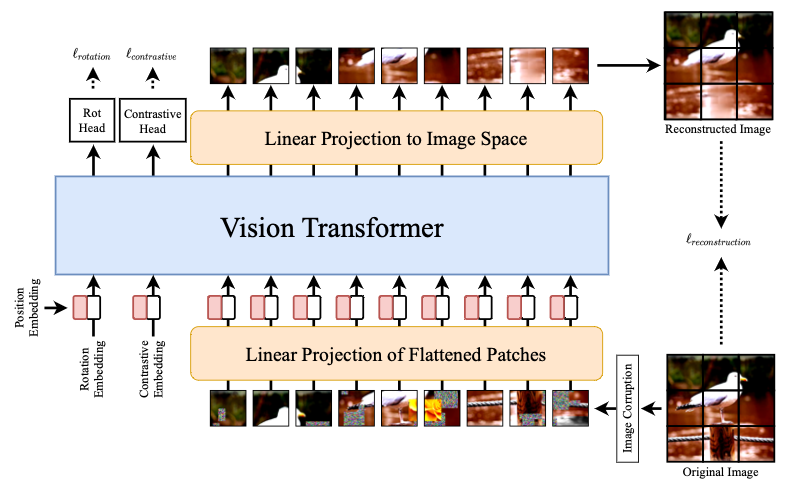}
   \caption{Architecture of SIT from \cite{Ahmed2021SiTSV}}
   \label{fig:sit}
\end{figure}

With the success of transformers in NLP, they are also getting increasingly popular in the computer vision community with the adapted vision transformers. Motivated by the performance gains of self-supervised methods in NLP, Self-supervised vIsion Transformer (SIT) \cite{Ahmed2021SiTSV} studied the effects of self-supervised learning for pretraining vision transformers and then used them for downstream classification tasks. As shown in Figure \ref{fig:sit}, the work used reconstruction of image patches as the main self-supervised task along with rotation prediction and contrastive learning to pre-train the vision transformer and then used them for downstream classification tasks. The results demonstrated the strength of the transformers with promising results and their suitability for self-supervised learning.

\begin{figure}[]
\centering
   \includegraphics[width=1\linewidth]{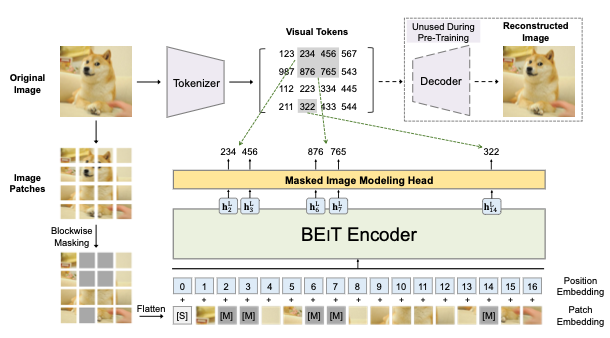}
   \caption{Architecture of BEiT from \cite{Bao2021BEiTBP}}
   \label{fig:beit}
\end{figure}

With the same motivation, BERT Pre-Training of Image Transformers (BEiT) \cite{Bao2021BEiTBP} directly adapted the Masked Language modeling from BERT to Vision Transformer with Masked Language modeling with minimal changes. In order to match textual tokens in BERT, BEiT used the pre-trained discrete Variational Autoencoder from \cite{Ramesh2021ZeroShotTG} to produce visual tokens as shown in Figure \ref{fig:beit}. Each image is represented in two views namely image patches and visual tokens during pretraining. Blockwise masking is applied over the input image and then the patches are fed to a backbone vision transformer. The objective of the pre-training task is to predict the visual tokens of the original image based on the encoding vectors of the masked image.

\begin{figure}[]
\centering
   \includegraphics[width=1\linewidth]{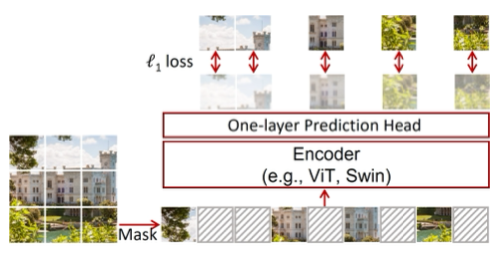}
   \caption{Architecture of SimMIM from \cite{Xie2021SimMIMAS}}
   \label{fig:sim-mim}
\end{figure}

The follow-up work, SimMIM \cite{Xie2021SimMIMAS}, removed the dependency on external visual tokenizer and proposed a simple framework for Masked Image modeling as shown in Figure \ref{fig:sim-mim}. It demonstrated powerful representation learning with the following design choices: 1) Using a moderately large patch size of 32 while masking the input for the Masked Image modeling task; 2) Using direct pixel value regression rather than image token classification 3) Using the shallow network in prediction head such as linear layer rather than a deep network. L1 loss is used as the reconstruction loss between the original and predicted pixel values that correspond to the masked patch positions.

Other kinds of transformation-based self-supervised learning methods have also been used to pre-train the neural networks. For example, \cite{Noroozi2016UnsupervisedLO} showed pre-training the convolution neural network with the task of predicting the correct positional index of the shuffled patches improved the network performance. However, the same is not shown for vision transformers. Also, another line-up of research works such as SimCLR\cite{Chen2020ASF}, Barlow Twins\cite{Zbontar2021BarlowTS} and VICReg\cite{Bardes2021VICRegVR} use contrastive loss for self-supervised pre-training neural networks.

Despite showing performing gains with the above shown developments of self-supervised learning methods in neural networks, it is not widely studied for the object detection task with vision transformer architectures. It could be because of the pre-trained CNN backbone that is commonly used to extract features from the input images before passing through the transformer layers. However, we view the CNN feature maps as the downsampled images and design transformation-based (non-contrastive) self-supervised methods to train the encoder in object detection transformers. Overall, we make the following contributions:
\begin{enumerate}
    \item We design 5 different self-supervised learning methods namely Image Reconstruction, MIM-Continuous, MIM-Discrete, Jigsaw-Continuous and Jigsaw-Discrete that use patch sizes in correspondence to the downsampling factor of the CNN backbone in detection transformer.
    \item We adapt 2 different learning strategies namely pre-training and multi-task learning to train the encoder in detection transformer with the above mentioned SSL methods.
    \item We demonstrate the improvements by SSL methods in DETR through preliminary experiments. However, our experiments show that the best performing SSL method in DETR i.e., reconstruction degrades the performance in case of Deformable DETR with multi-task learning. 
    \item We propose to utilize our self-supervised pre-training approach with a massive number of unlabelled images to efficiently train the object detection transformer with limited annotation budget in the real-world.
\end{enumerate}

\section{Related Work}
\label{sec:rel}
\subsection{Object Detection Transformers}
Traditional object detectors are usually convolution-based with either single or two-stage models incorporating a lot of hand-crafted anchors. However, transformers have gained significant interest and are the focus of study in our work. Object detection methods have been utilizing the self-attention mechanism of transformers and then enhancing on top of it with specific modules on generalized scope or a fixed set. 

\subsection*{DETR}

\begin{figure*}[]
\centering
   \includegraphics[width=1\linewidth]{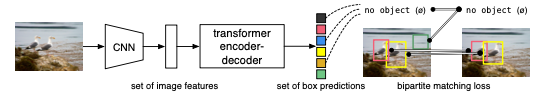}
   \caption{Overview of DETR from \cite{carion2020end}}
   \label{fig:detr}
\end{figure*}

DEtection TRansformer (DETR)\cite{carion2020end} proposed a transformer-based end-to-end object detector without the inclusion of handcrafted components like anchor design or Non-Maximal Suppression for detection that explicitly encodes the prior knowledge, as shown in Figure \cite{fig:detr}. the major contribution was a global bipartite loss which enforced unique matchings with a multi-layer transformer encoder-decoder framework with multiple prediction heads. The conceptualized model was simple and did not require any specialized backbones while demonstrating performance on par with the well-developed Faster R-CNN \cite{ren2015faster} on the arduous COCO detection task as it leverages learnable queries to check on the existence of objects in combination with feature map. The biggest improvement presented came in the detection of large objects because of the global information being b=processed by self-attention. Despite the simple design and great improvements in performance, the training convergence of DETR is pretty slow and the meaning of queries on smaller objects is unclear.

\subsection*{Deformable DETR}

\begin{figure}[]
\centering
   \includegraphics[width=1\linewidth]{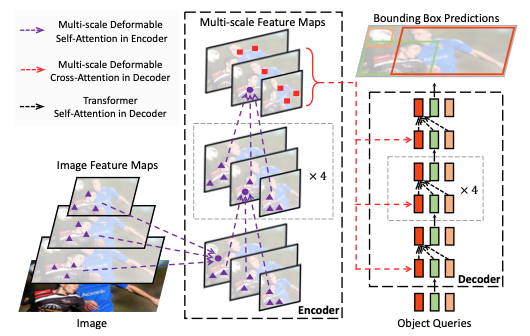}
   \caption{Architecture of Deformable DETR from \cite{dai2017deformable}}
   \label{fig:ddetr}
\end{figure}

To address multiple problems in DETR, Deformable DETR (Figure \ref{fig:ddetr}) introduces deformable attention by bringing in the best of sparse spatial sampling from deformable convolutions and relation modeling. The proposed deformable attention module tends to look for important key elements from all feature maps for a smaller sample. It explored a simple iterative bounding box refinement without the help of FPN \cite{lin2017feature}. It showed significantly faster convergence (up to 10x) with computational and memory efficiency. Inspired by deformable convolutions \cite{dai2017deformable}, it attends to a small set of points from a reference and hence by assigning only a small number of keys for each query

\subsection*{Other Object Detection Transformers}
There have been several directions of improvement over DETR. Bridging between transformers and classical anchor-based model,  DAB-DETR \cite{liu2022dab} proposed to formulate DETR queries as dynamic anchor boxes. It extends 2D anchor points to 4D anchor box coordinates to represent queries. The bipartite matching process is somewhat unstable, which is solved with additional denoising in DN-DETR \cite{li2022dn}. Further DINO-DETR \cite{zhang2022dino} formulates dynamic queries like DAB-DETR and then refines them across decoder layers. To improve denoising DINO-DETR introduces a contrastive denoising training by adding both positive and negative samples simultaneously, hence making a mixed query selection method and a \textit{look forward twice} scheme to better update parameters from gradients from later layers while still adopting deformable attention for computational efficiency.

% \textcolor{red}{@abhishek add 2 paras for Cell-DETR, XCI \& YOLOS}

Cell-DETR \cite{prangemeier2020c} presented another attention-based approach with transformers for direct instance segmentation for a particular application in biomedical data, cited as one of the first transformer-based detection work in the field. They put forth two variants of the work, a regular one that matches the performance of Mask R-CNN while using lesser parameters and 70\% of the time; and a lighter one which needs less than a third of the time from Mask R-CNN with about only 1 point drop in performance. However it was developed on and for a specific task of cell segmentation in clusters, it was adaptable to several biomedical applications while achieving state-of-the-art performance.

XCiT (Cross-Covariance Image Transformers) \cite{ali2021xcit} is an alternative to self-attention working on tokens only, as it replaces compute expensive quadratic attention with operations on feature dimension. They introduced a transposed form of attention, which they term "cross-covariance attention (XCA)" to substitute the full pairwise interaction among tokens with a more efficient attention mapping. It showed a strong performance on par with state-of-the-art works in image classification while staying invariant to scale changes. It was also proposed as a strong backbone for dense prediction tasks for detection and segmentation with self-supervised learning that eases the compute for better results. The transferability of ViT pre-training on lesser sized datasets for application on complex problems was explored with YOLOS \cite{Fang2021YouOL}. It experimented with object detection based on a sequence-to-sequence manner with very few inductive biases added. This supports the versatility and adaptability of transformers in downstream tasks.

% - DN Detr
% - Dino Detr
% - YOLOS
% - Cell-Detr
% - XCiT: Cross-Covariance Image Transformers

\subsection{Self-Supervised Learning in Object Detection Transformers}

%\textcolor{red}{@sahal add 2 paras for UP-DETR and Det-Reg}
To address the training and optimization challenges of DETR, like longer training times and large-scale training data UP-DETR \cite{UP-DETR} was introduced. By introducing a novel pretext task named \textit{random query patch detection} UP-DETR outperforms DETR with faster convergence and better performance. DETR has a CNN backbone which is pre-trained to extract low-level image representations. UP-DETR proposes to first pre-train the Transformer in a self-supervised manner and then fine-tune the entire model with labeled data. The main idea of the random query patch detection is to randomly crop patches from the given image and then feed them as queries to the decoder. The transformer is pre-trained in a self-supervised manner to detect the location of these cropped patches. The pretext task is done in multi-task learning and multi-query localization settings and the proposed approach improves the performance for downstream tasks like object detection, one-shot detection and panoptic segmentation. 

DETReg \cite{DETReg} introduces a novel self-supervised method that pre-trains the entire object detection network and proposes to improve the transformer-based object detection performance in low-data regimes. DETReg (DEtection with TRansformers based on Region priors) uses two key pretraining tasks to train a detector on unlabelled data, namely - \textit{Object Localization Task} and \textit{Object Embedding Task}. Object Localization task aims to train the model to localize objects, regardless of their categories. While the Object Embedding task is geared toward understanding the categories of objects in the image. DETReg is built on top of the Deformable DETR architecture which offers faster convergence and simplified implementation.

\section{Methodology}
\label{sec:method}

\subsection{Self-Supervised Learning Tasks}

\begin{figure*}[t]
\centering
   \includegraphics[width=1\linewidth]{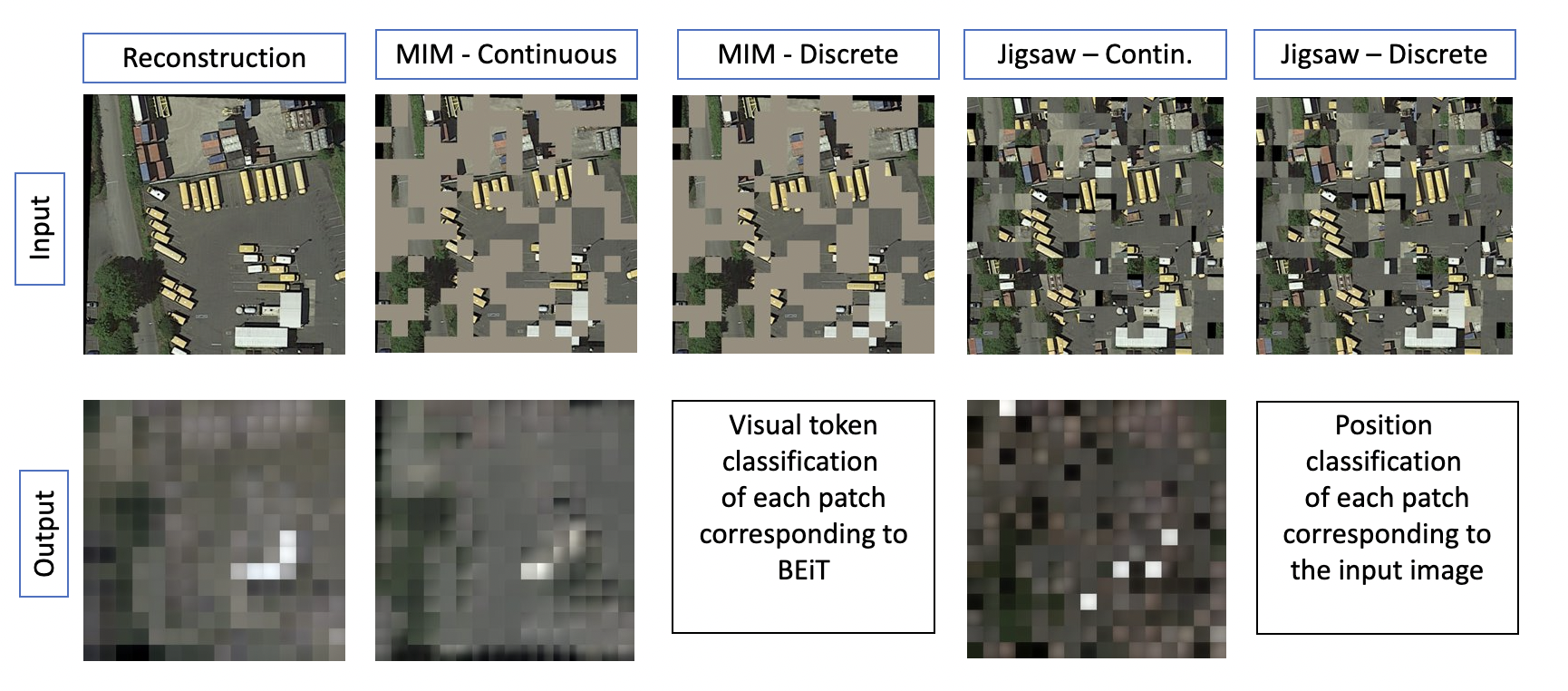}
   \caption{Input and output of different self-supervised methods studied in our work corresponding to iSAID images}
   \label{fig:ssl}
\end{figure*}

The input and output representations for all of our self-supervised methods are shown in Figure \ref{fig:ssl}.

\textbf{Reconstruction:} The unmodified original image is given as the input. The L1 loss between the predicted image and original image is used to optimize the encoder.

\textbf{MIM-Continuous:} The original image is divided into patches conditioned by the CNN backbone's downsampling factor. A specific number of random patches are masked with average pixel values based on the SSL task ratio and given as the input. The L1 loss between the predicted image and original image, corresponding only to the masked patches, is used to optimize the encoder.

\textbf{MIM-Discrete:} The patch formation and masking are same as MIM-Continuous. The cross-entropy loss between the actual visual token and the predicted visual token, corresponding only to the masked patches, is used to optimize the encoder. In our experiments, we found that the discrete Variational Autoencoder used in BEiT is computationally intensive and time consuming for tokenization. Hence, we did not evaluate this method.

\textbf{Jigsaw-Continuous:} The original image is divided into patches conditioned by the CNN backbone's downsampling factor. A specific number of random patches are selected and permuted based on the SSL task ratio and given as the input. The L1 loss between the predicted image and original image, corresponding only to the permuted patches, is used to optimize the encoder.

\textbf{Jigsaw-Discrete:} The patch formation and masking are same as Jigsaw-Continuous. The cross-entropy loss between the actual patch position and the predicted patch position, corresponding only to the permuted patches, is used to optimize the encoder.

\subsection{Encoder Training with SSL}
\begin{figure*}[]
\centering
   \includegraphics[width=1\linewidth]{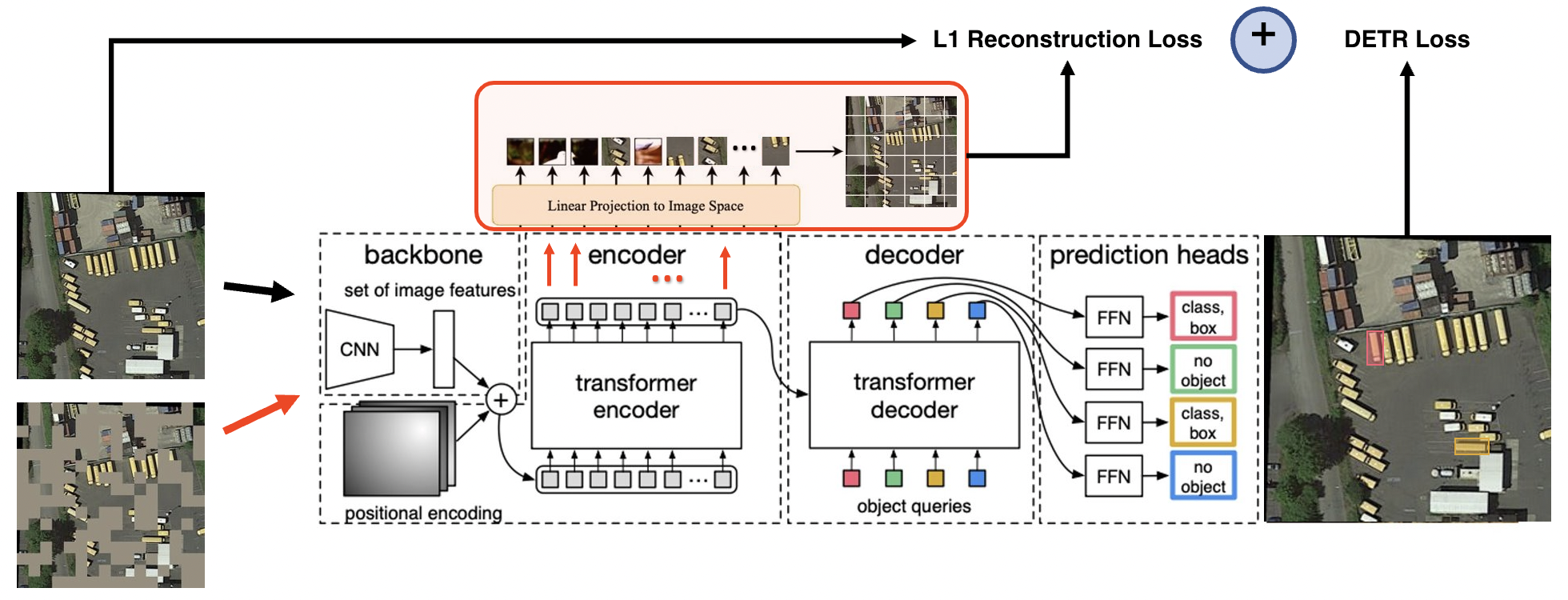}
   \caption{Architecture of DETR with self-supervised learning task head (red block). Red arrows indicate its corresponding data flow.}
   \label{fig:detr-with-ssl}
\end{figure*}

\textbf{Multi-task Learning:} In Figure \ref{fig:detr-with-ssl}, we illustrate the addition of a self-supervised learning task head with the DETR encoder in a multi-task learning setup. The transformed input image corresponding to the given SSL method is passed through the CNN backbone, added with positional encoding, passed through the transformer encoder block and finally through the SSL task head to make the prediction. The resulting SSL loss is combined with the original DETR loss that is computed by passing the unmodified original image in parallel. This total loss is used to optimize the network. The weight of the SSL loss could be decreased over time for effective learning. We follow the same approach to add the SSL task head in Deformable DETR by computing the SSL loss only at the maximum image scale.

\textbf{Pre-training:} At first, only the SSL task loss is optimized using the training images. The pre-training dataset need not be labeled and is not strictly required to be from the same distribution as the target fine-tuning dataset. Then, DETR or Deformable Loss is used to optimize the whole network with the target dataset.

\section{Experiments}
\label{sec:exp}

\subsection{Datasets}
\textbf{ImageNet:} The ImageNet dataset is an extremely popular collection of images that are commonly used as a pre-training dataset for many downstream image-based tasks. It originally consists of over 3.2 million images from various categories organized according to WordNet \cite{dataset_imagenet}. The images in ImageNet are typically observed in the wild, either against a natural background or wherever these objects typically reside. 

\textbf{COCO:} The COCO dataset, also known as Common Objects in Context is a similar dataset to Imagenet with objects in the location where they typically reside. However, unlike ImageNet's thousands of categories, COCO contains 91 object categories \cite{dataset_coco} but the data only uses 80 classes. The dataset has 1.5 million object instances and could be used for object detection, image segmentation and image captioning. Despite having multiple object detection datasets, COCO is widely used as the standard benchmark for the evaluation of object detection models in the computer vision community. The COCO model weights contribute to an effective pre-training method and improve the final AP score as the model has learned the downstream task of Object Detection. 

\begin{figure}[]
\centering
   \includegraphics[width=1\linewidth]{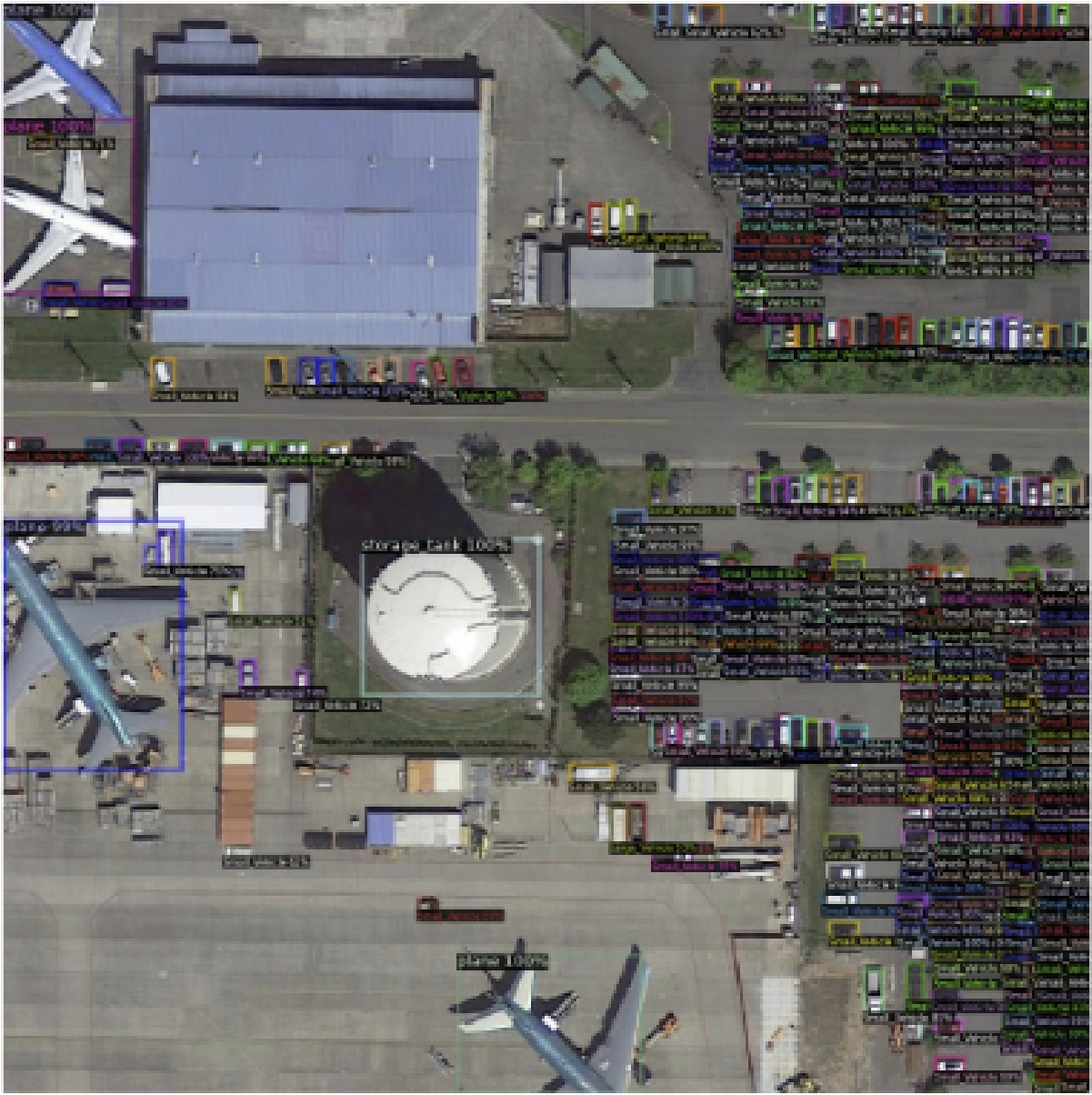}
   \caption{Sample aerial image from the iSAID data with bounding box annotations showing small and dense bounding boxes.}
   \label{fig:isaid}
\end{figure}

\textbf{iSAID:} It is a large-scale dataset for Instance segmentation in aerial images \cite{waqas2019isaid} . The dataset contains a total of 2,806 high-resolution images with 655,451 object instances for 15 categories. A sample image is shown in Figure \ref{fig:isaid}. The iSAID dataset is the first benchmark dataset for instance segmentation for satellite images and is used for both Object Detection and Instance segmentation. The dataset was created by annotating the original DOTA \cite{Xia_2018_CVPR} dataset with 2806 images and it contains roughly 3.5 times more number of instances than DOTA.

\subsection{Setup}
We fine-tune our models with the train split and evaluate the dev and test splits of the iSAID dataset. For pre-training, we use ImageNet and MSCOCO datasets. We use Pytorch on NVIDIA Tesla A100 GPU, with 40 GB dedicated memory, with CUDA-11.1 installed. The hyper-parameter values for all DETR models  are shown in Table [\ref{tab:hyper}]. For DETR models in pre-training setup, we pre-train the model initially for 10 epochs using ImageNet and fine-tune for 100 epochs. For DETR models in multi-task learning setup, we fine-tune the MSCOCO pre-trained model for 50 epochs. For Deformable DETR we set the batch size to 8. In all these, the ratio of the self-supervised task is set as 0.5

\begin{table}[!h]
\begin{center}
\begin{tabular}{c c}

\hline
Hyperparameter        & Value \\
\hline
Minimum Image size & 512 \\
Maximum Image size & 512 \\
Number Object Proposals & 100 \\
Optimizer & AdamW \\
Batch size & 64 \\
Learning rate & 0.0001 \\
Weight decay & 0.0001 \\
Gradient clip value & 0.1 \\
\hline
\end{tabular}
\end{center}
\caption{Hyperparameter configuration of our models.}
\label{tab:hyper}
\end{table}

\subsection{Evaluation Metrics}
Standard COCO evaluation metrics were used to compare and analyze the different models. The COCO evaluation metric uses a range of IoU (Intersection over Union) thresholds from 0.5 to 0.95 with a step size of 0.05 represented as AP@[.5:.05:.95]. The Average Precision (AP) for thresholds 0.5-0.95 is calculated and then averaged to get the mean Average Precision (mAP). This is done in order to reduce bias and give equal representation to all threshold values.
\begin{equation}
\small
    \label{eq:map}
    mAP_{COCO} = \frac{mAP_{0.50}+mAP_{0.55}+...+mAP_{0.95}}{10} \text{\cite{coco_eval}}
\end{equation}
The COCO evaluation metric makes no distinction between Average Precision (AP) and mean Average Precision (mAP) and it is calculated using Equation [\ref{eq:map}]. AP50 and AP75 stand for AP calculated at IoU = 50 \& 75 respectively. The Average Precision for small, medium and large objects is given by APs, APm, APl respectively.

\subsection{Results}

\begin{figure}[]
\centering
   \includegraphics[width=1\linewidth]{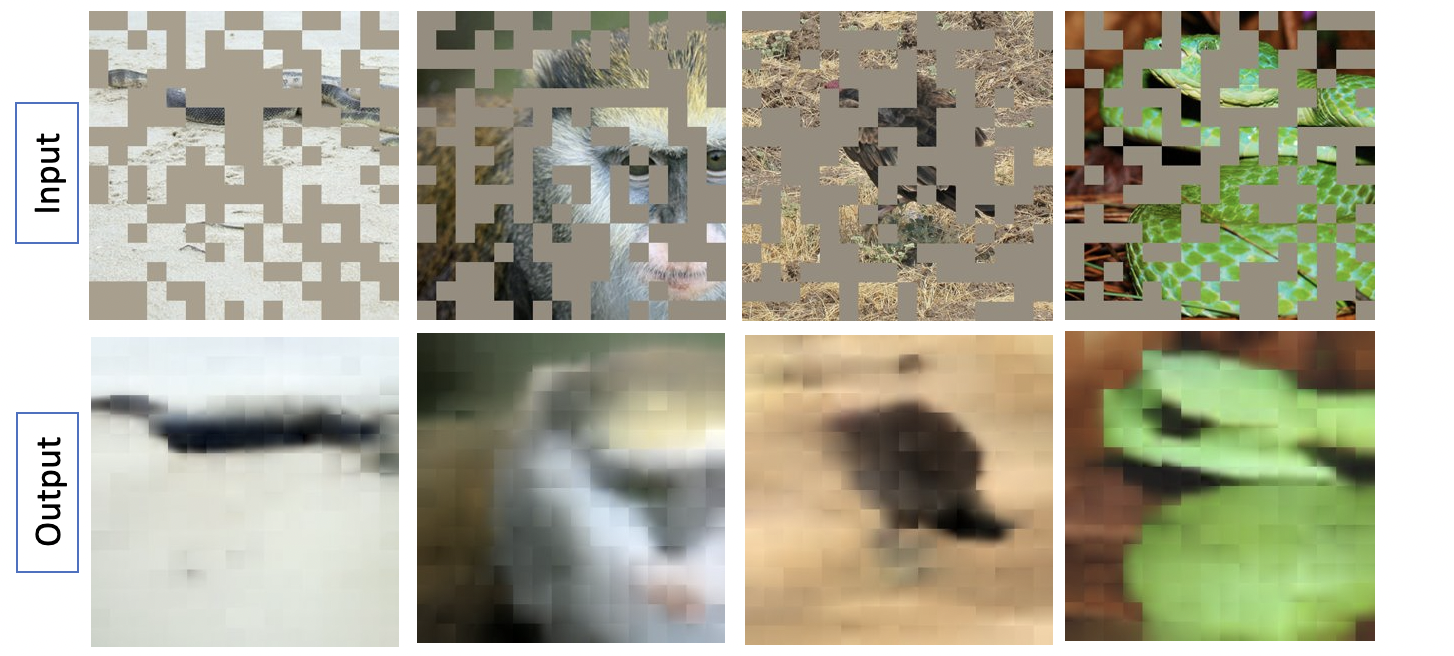}
   \caption{Masked input images and their corresponding output images predicted by the DETR encoder after self-supervised pretraining with Masked Image modeling task on the ImageNet dataset.}
   \label{fig:mim-imagenet}
\end{figure}

\begin{table*}[]
\begin{center}
 \resizebox{0.9\textwidth}{!}{%
\begin{tabular}{|c|c|c|c|c|c|c|c|c|}
\hline
\textbf{SSL Task} &  \textbf{SSL Task Ratio} & \textbf{AP} & \textbf{AP$_{50}$}& \textbf{AP$_{75}$}& \textbf{AP$_S$} & \textbf{AP$_M$} & \textbf{AP$_L$} & \textbf{Training hours} \\ \hline
Baseline: DETR & - & 0.433 & 1.078 & 0.259 & 0.128 & 0.443 & 1.083 & 20.5 \\ \hline
+ MIM-Continuous & 0.5 &0.734 & 1.894 & 0.495  & 0.263 & 0.491 & 1.963 & 26.5 \\ \hline
+ Jigsaw-Continuous & 0.5 & 1.317 & 3.680 & 0.755  & 0.768 & 1.453 & 2.938 & 19.9 \\ \hline
+ Jigsaw-Discrete & 0.5 & 0.367 & 1.058 & 0.133  & 0.228 & 0.358 & 0.514 & 19.3 \\ \hline

\end{tabular}%
}   
\end{center}
\caption{Development set performance of different SSL methods with DETR in pretraining setting after 100 epochs.}
\label{tab:detr-img}
\end{table*}

\begin{table*}[]
\begin{center}
 \resizebox{0.9\textwidth}{!}{%
\begin{tabular}{|c|c|c|c|c|c|c|c|c|}
\hline
\textbf{SSL Task} &  \textbf{SSL Task Ratio} & \textbf{AP} & \textbf{AP$_{50}$}& \textbf{AP$_{75}$} & \textbf{AP$_S$} & \textbf{AP$_M$} & \textbf{AP$_L$} & \textbf{Training hours} \\ \hline
Baseline: DETR  & - & 6.954 & 16.480 & 6.343 & 1.903 & 7.923 & 17.680 & 18.2 \\ \hline
+ Reconstruction & - & 8.954 & 20.050 & 7.270 & 2.748 & 10.080 & 21.060 & 15.5 \\ \hline
+ MIM-Continuous & 0.5 & 8.171 & 18.340 & 6.820 & 2.147 & 8.549 & 17.430 & 15.5 \\ \hline
+ Jigsaw-Continuous & 0.5 & 7.551 & 16.830 & 6.543 & 1.814 & 7.123 & 15.760 & 18.5 \\ \hline
+ Jigsaw-Discrete & 0.5 & 7.369 & 16.970 & 4.885 & 1.850 & 7.440 & 16.690 & 16.1 \\ \hline
\end{tabular}%
}   
\end{center}
\caption{Development set performance of different SSL methods with DETR in multi-task learning setting after 50 epochs.}
\label{tab:detr-mtl}
\end{table*}

\begin{table*}[!h]
\begin{center}
 \resizebox{0.9\textwidth}{!}{%
\begin{tabular}{|c|c|c|c|c|c|c|c|c|}
\hline
\textbf{SSL Task} & \textbf{SSL Task Ratio} & \textbf{Epochs} & \textbf{AP} & \textbf{AP$_{50}$} & \textbf{AP$_{75}$} & \textbf{AP$_S$} & \textbf{AP$_M$} & \textbf{AP$_L$} \\ \hline
Baseline: Deformable DETR & - & 100 & 41.0 & 61.6 & 45.3 & 43.2 & 50.6 & 14.0 \\ \hline
Baseline: Deformable DETR & - & 25 & 29.2 & 48.6 & 31.1 & 31.1 & 33.1 & 8.7  \\ \hline
+ Reconstruction & 0.5 & 25 & 27.6 & 47.2 & 28.9 & 28.7 & 35.3 & 8.0 \\ \hline
\end{tabular}%
}   
\end{center}
\caption{Test set performance of SSL method with Deformable DETR in multi-task learning setting.}
\label{tab:ddter}
\end{table*}

For the first set of experiments, the DETR model was pre-trained on the ImageNet dataset for 10 epochs. The input and predicted images of the self-supervised pretraining task in ImageNet dataset, with masked image modeling of task ratio 0.5, are shown in Figure \ref{fig:mim-imagenet}. We further fine-tune the models on the given iSAID dataset for 100 epochs.  The overall AP score for these models are very low (See Table \ref{tab:detr-img}). DETR models usually require 500+ epochs to converge as DETR has a Quadratic dependency on the input sequence length, hence due to finite computational resources, our experiments were limited to 100 epochs. Nevertheless, the transformer encoder pre-training and reconstruction tasks show improvement over the baseline. Jigsaw Continuous and MIM Continuous increase the baseline score from 0.433 AP to 1.317 and 0.734 AP. However, Jigsaw Discrete seems to adversely affect the performance. 

For the next set of experiments, the pre-trained DETR model with MSCOCO weights was used in a multi-task learning setting. The results of these experiments are considerably better compared to Table \ref{tab:detr-img} as the DETR model was pre-trained on MSCOCO for 500 epochs and it has learned the downstream task of object detection. We further fine-tune these models on the iSAID dataset for 50 epochs with Self-Supervised learning approaches. In this setting, all the models show improvement over the baseline. The MIM Continuous with SSL task ratio set to 0 gives the best model with an AP score of 8.954 (See Table \ref{tab:detr-mtl}). We can see from Tables \ref{tab:detr-img} and \ref{tab:detr-mtl} that the AP score for Large objects are considerably higher than the small and medium categories. This can be attributed to the fact that DETR does not use any FPN (Feature Pyramid Network) to make predictions at different scales and also it uses low-resolution down-sampled feature map from its CNN backbone.

Finally, Deformable DETR (DDETR) model was used to solve the convergence and small object AP issue. DDETR shows significant improvement in the AP score compared to DETR in Table \ref{tab:detr-img} and Table \ref{tab:detr-mtl}. The baseline DDETR ran for 100 epochs gives an overall AP score of 41.0 and 43.2 AP for small objects (See Table \ref{tab:ddter}). Large AP performance is poor when compared to other categories, this might be because the Deformable Attention attends only to a small set of points. The SSL reconstruction model was run for only 25 epochs due to limited computational resources and time constraints. However, this model performance (27.6) was inferior to the baseline DDETR (29.2) which ran for 25 epochs. The Deformable Attention might be adversely affecting the reconstruction tasks and the model needs to be run for 100+ epochs to converge and further validate our hypothesis.

\section{Real World Application}
While developing the objection detection model for aerial images, we also studied how different companies utilize aerial images in the real world. We approached one such promising startup called Smart Navigation\footnote{\url{http://smartnavigation.ae/}}, which has massive aerial imaging datasets and wants to develop an object detection model for aerial images as part of their upcoming product Eagle Eye. However, the budget is sufficient only to annotate a fraction of their acquired images. In this case, all the unlabeled images could be utilized with our proposed SSL approaches in pre-training setup of object detection transformers.

\section{Conclusion}

% \textcolor{red}{@abhishek Add 1 or 2 para for conclusion}

% self-supervised learning (SSL) methods such as masked language modeling have shown massive performance gains by pretraining transformer models for a variety of natural language processing tasks. The follow-up research adapted similar methods like masked image modeling in vision transformer and demonstrated improvements in the image classification task. Such simple self-supervised methods are not exhaustively studied for object detection transformers (DETR, Deformable DETR) as their transformer encoder modules takes input in the convolutional neural network (CNN) extracted feature space rather than the image space as in general vision transformers. However, the CNN feature maps still maintain the spatial relationship and we utilize this property to design self-supervised learning approaches to train the encoder of  object detection transformers in pretraining and multi-task learning settings. We explore common self-supervised methods based on image reconstruction, masked image modeling and jigsaw. Preliminary experiments in the iSAID dataset demonstrate faster convergence of DETR in the initial epochs in both pretraining and multi-task learning settings; nonetheless, similar improvement is not observed in case of multi-task learning with Deformable DETR.

We motivated the need for incorporating SSL techniques in the object detection transformer (DETR/Deformable-DETR) to train its encoder effectively. Then, we designed and experimented 5 different SSL methods, based on the feature maps produced by the pretrained CNN backbone of the object detection transformer, in two settings namely: pretraining and multi-task learning. Our preliminary experiments with the iSAID dataset demonstrate a faster convergence with DETR for both pre-training and multi-task learning settings in the intial epochs. However, similar improvement is not seen in the case of multi-task learning with Deformable DETR which still improves over DETR in terms of overall AP and in terms of medium objects. We further illustrated the real world use case of our proposed technique in case of dealing with a huge number of non-annotated satellite images.

This work is done as a final project for our graduate course "Visual Object Detection and Recognition" (CV703) at MBZUAI. Due to the computational and time constraints, only a subset of the planned experiments were executed and studied in limited settings. In order to holistically validate our proposed techniques, more comprehensive evaluations have to be done based on this work. Some of the possible extensions are listed as follows: (a) DETR models have to be trained 10X times or until convergence, (b) all SSL methods and settings have to be evaluated with Deformable-DETR, and (c) Decaying SSL loss weight / SSL task ratio in the multi-task learning setup.

 \newpage

% \textcolor{red}{@abhishek Check grammar, spelling, references and formatting}

% \textcolor{red}{@sahal Record teaser with some of our presentation slides}

\label{sec:con}

%%%%%%%%% REFERENCES
{\small
\bibliography{references}{}
\bibliographystyle{unsrt}
}

\end{document}